%% file: main.tex
\documentclass[10pt,twocolumn,letterpaper]{article}

\usepackage[pagenumbers]{cvpr} 
\usepackage[accsupp]{axessibility}

\input{preamble}

\definecolor{cvprblue}{rgb}{0.21,0.49,0.74}
\usepackage[pagebackref,breaklinks,colorlinks,citecolor=cvprblue]{hyperref}

\usepackage{makecell, booktabs}
\usepackage{ulem}

\usepackage{animate}
\newlength{\itemwidth}

\def\paperID{7311} 
\def\confName{CVPR}
\def\confYear{2024}

\title{StyleCineGAN: Landscape Cinemagraph Generation\\
using a Pre-trained StyleGAN}

\author{
Jongwoo Choi \qquad Kwanggyoon Seo \qquad Amirsaman Ashtari \qquad Junyong Noh \\
    Visual Media Lab, KAIST \\ 
{\small{\url{https://jeolpyeoni.github.io/stylecinegan_project/}}}
}

\begin{document}


\twocolumn[{
\maketitle

\begin{center}

    \vspace{-0.5cm}

    \setlength{\tabcolsep}{0.12cm}
    \setlength{\itemwidth}{5.2cm}
    
    \captionsetup{type=figure}
    \begin{tabular}{ccc}
        \animategraphics[width=\itemwidth, autoplay, final, loop, nomouse, method=widget, poster=last]{24}{graphics/teaser_small/0000267/}{000}{047} &
        \animategraphics[width=\itemwidth, autoplay, final, loop, nomouse, method=widget, poster=last]{24}{graphics/teaser_small/0003835/}{000}{047} &
        \animategraphics[width=\itemwidth, autoplay, final, loop, nomouse, method=widget, poster=last]{24}{graphics/teaser_small/0001369_blue/}{000}{047}
         \\
    \end{tabular}
    \vspace{-0.28cm}
    \captionof{figure}{{Given a landscape image, StyleCineGAN generates a seamless cinemagraph at 1024$\times$1024 resolution.~\textbf{This figure contains video clips, thus consider viewing it using Adobe Reader}.~The same results are also~included~in~the supplementary~video.}}

    \label{fig:teaser}

\end{center}

}]

\newcommand{\kg}[1]{\textcolor{blue}{[\textbf{kg}:~#1]}}
\newcommand{\jw}[1]{\textcolor{red}{[\textbf{jw}:~#1]}}
\newcommand{\sam}[1]{\textcolor{cyan}{[\textbf{sam}:~#1]}}
\newcommand{\revis}[1]{\textcolor{blue}{#1}}
\newcommand{\spot}[1]{\textcolor{red}{#1}}


\input{source/100_abstract}
\input{source/100_intro}

\input{source/200_related}

\input{source/400_method}
\input{source/600_experiment}

\input{source/700_discussion}

\input{source/800_conclusion}



{
    \small
    \bibliographystyle{ieeenat_fullname}
    \bibliography{main.bib}
}

\end{document}

%% file: preamble.tex
\usepackage[dvipsnames]{xcolor}


%% file: source/100_abstract.tex
\begin{abstract}

\vspace{-0.35cm}

We propose a method that can generate cinemagraphs automatically from a still landscape image using a pre-trained StyleGAN. Inspired by the success of recent unconditional video generation, we leverage a powerful pre-trained image generator to synthesize high-quality cinemagraphs. Unlike previous approaches that mainly utilize the latent space of a pre-trained StyleGAN, our approach utilizes its deep feature space for both GAN inversion and cinemagraph generation. Specifically, we propose multi-scale deep feature warping (MSDFW), which warps the intermediate features of a pre-trained StyleGAN at different resolutions. By using MSDFW, the generated cinemagraphs are of high resolution and exhibit plausible looping animation.
We demonstrate the superiority of our method through user studies and quantitative comparisons with state-of-the-art cinemagraph generation methods and a video generation method that uses a pre-trained StyleGAN.


\vspace{-0.2cm}

\end{abstract}

%% file: source/100_intro.tex
\vspace{-0.3cm}
\section{Introduction}

\vspace{-0.1cm}


Cinemagraph is a unique form of media that combines a still image and video. While most of the scene remains still, subtle and repeated movements in only a small region effectively highlight a particular object or an important event that occurred at the time of capture. Because this mixture of still and moving elements in one scene creates an interesting eye-catching effect, cinemagraphs have recently attracted increasing attention in social media.

Despite the growing popularity of cinemagraph, its creation heavily relies on a manual process. To create a motion in part of the input image, one typically utilizes an image manipulation tool to stretch, scale, rotate, or shift the part of the image in a physically plausible and aesthetically pleasing way. This process is often time-consuming and requires a high level of skill in the use of image editing tools. Therefore, creating a cinemagraph has generally been  considered to be a personal project of professionals so far.


To allow ordinary users to create a cinemagraph with a single image, automatic methods have been proposed. One line of research uses a reference video as guidance~\cite{prashnani2017phase, tesfaldet2018, LoopNeRF}. These methods are capable of producing realistic motions in the target scene similar to a reference video.
Recent methods obviate the need for a reference video, by training deep generative models~\cite{animating-landscape, Logacheva_2020_ECCV, Holynski_2021_CVPR, controllable, text2cinemagraph, li2023_3dcinemagraphy, bertiche2023blowing, pajouheshgar2022dynca}. These methods decompose the task into two processes of learning motion and spatial information individually from separate datasets. The decomposition effectively reduces the complexity of simultaneously learning both temporal and spatial information, and improves the perceptual quality of the generated videos. However, these models must be trained from scratch, which sometimes requires several days if not weeks on a modern GPU.
Moreover, these models~are~not~specifically~designed to~generate~high-resolution~of~1024$\times$1024~cinemagraphs, because of the significant requirements in memory~and~processing~power.



\vspace{-0.03cm}

In this paper, we propose the first approach to high-quality one-shot landscape cinemagraph generation based on a pre-trained StyleGAN~\cite{stylegan2}. By using the learned image prior of StyleGAN, our method removes the need for training a large model from scratch and systematically improves the resolution of the generated cinemagraphs to 1024$\times$1024. Moreover, our method enables a user to easily edit the style and appearance of the generated cinemagraph by leveraging the properties of StyleGAN for image stylization~and~editing.

\vspace{-0.03cm}

Our method is inspired by recent unconditional video generation methods~\cite{fox2021stylevideogan, tian2021a} which allow to navigate the latent space of a pre-trained image generator, to synthesize a high-quality temporally coherent video.
Unlike these methods that utilize the latent codes of StyleGAN, we opt to use the deep features that are generated by convolution operations in each layer of StyleGAN. We use these deep features for two reasons.
First, we observed that highly detailed landscape images cannot be reconstructed accurately from the latent codes using GAN inversion methods because these latent codes are low-dimensional~\cite{rate-distortion}.
Second, a plausible motion that preserves the content cannot be created by only navigating the latent space, because this space is highly semantic-condensed and lacks explicit spatial prior~\cite{wang2021HFGI}.

\vspace{-0.03cm}

To solve these problems, for the first time in the cinemagraph generation domain, we utilize the deep features of a pre-trained StyleGAN. These deep features preserve spatial information and encode both high-level semantic and low-level style appearances across high and low-level convolutional layers. To produce cinemagraphs from these deep features, we propose a multi-scale deep feature warping (MSDFW) to apply the motions generated using a motion generator to the deep feature space of StyleGAN.



\vspace{-0.03cm}

We demonstrate the effectiveness and advantages of our method by comparing it to state-of-the-art methods in cinemagraph generation~\cite{animating-landscape, Logacheva_2020_ECCV, Holynski_2021_CVPR, text2cinemagraph} and to an unconditional video generation method that uses a pre-trained StyleGAN~\cite{tian2021a}, using various metrics. We also performed a user study to verify the effectiveness of our method for creating visually pleasing cinemagraph effects. Both qualitative and quantitative results confirm that our method substantially outperforms all existing baselines.

\vspace{-0.1cm}

%% file: source/200_related.tex
\section{Related Work}

\subsection{Cinemagraph Generation}
One of the early studies for cinemagraph generation used a procedural approach to decompose each object into layers and apply time-varying stochastic motion~\cite{chuang2005animating}.
This method can handle a diverse range of images such as photos and paintings by relying on manual interaction from the user. Another approach is to use a reference video to animate the given image~\cite{okabe2009animating, okabe2011creating, okabe2018animating, prashnani2017phase}. These methods rely on a statistical motion analysis of videos to transfer their periodic motion property to the desired image. Another method distills a dynamic NeRF to render looping 3D video textures from the representative motion found in the reference video~\cite{LoopNeRF}. Without any reference video as guidance, \citet{halperin2021endless} proposed a framework to animate arbitrary objects with periodic patterns in a simple motion direction. 

Recent approaches use the capacity of deep learning to automatically create cinemagraphs from a single image. One study trained an image-based renderer 
to generate water animation, utilizing the water simulation results~\cite{WaterSimulRendering}.
Another research first predicts a sequence of normal maps, then generates the corresponding RGB images that show a looping animation of a garment as if it is blown in the wind~\cite{bertiche2023blowing}.
Another line of research trains~a generator to produce a motion field~\cite{animating-landscape,Holynski_2021_CVPR,controllable,text2cinemagraph,li2023_3dcinemagraphy}.
Our work is similar to these deep learning-based approaches in that a motion generator is utilized to synthesize the cinemagraphs. The difference is that our method leverages the deep features of a pre-trained image generator and thus can generate plausible high-quality results without fine-tuning or training a separate synthesis network. At the same time, our method systematically improves the resolution of the generated cinemagraphs to 1024$\times$1024.

\vspace{-0.1cm} 
\subsection{Unconditional Video Generation}
\vspace{-0.1cm}
The task of video generation is known to be difficult because the generation process has to take into account both spatial and temporal information.
To ease the problem, previous approaches~\cite{Tulyakov:2018:MoCoGAN,saito2020tgan} decompose video generation into content and motion generation. These methods first predict the latent trajectory for motion, then generate a video from the set of predicted latent codes using the image generator. Instead of training the generation model from scratch, MoCoGAN-HD~\cite{tian2021a} and StyleVideoGAN~\cite{fox2021stylevideogan} leverage a pre-trained image generator model, StyleGAN~\cite{stylegan2}. These methods use the property of well-constructed latent space to morph one frame toward the next frames~\cite{gansteerability}. Utilizing the image generation capability of a pre-trained generator, MoCoGAN-HD and StyleVideoGAN only need to train a motion trajectory generator, which greatly reduces the training time. 
These methods can generate high-quality videos in multiple domains such as faces, cars, and even sky.
However, the content details are not well preserved, which has hindered the methods' application for cinemagraph generation.
Our method builds upon a similar concept of using a pre-trained StyleGAN. We leverage the deep features instead of the latent codes of StyleGAN to generate looping and content-preserving cinemagraphs.


\subsection{Video Synthesis using pre-trained StyleGAN}
A pre-trained generative model, StyleGAN~\cite{stylegan1,stylegan2,stylegan3}, has been actively employed for downstream image and video editing applications. For GAN-based applications, one of the essential steps is to project the desired images or videos to the latent space of the pre-trained GAN model, which is known as GAN inversion. The latent codes are obtained using either an optimization technique~\cite{abdal2019image2stylegan, abdal2019image2styleganpp}, a trained encoder~\cite{richardson2021psp,tov2021e4e}, or a hybrid of both approaches~\cite{lin2021anycost}. For video editing applications, previous methods~\cite{alaluf2022third, seo2022spv, tzaban2022siit, xu2022temporally} have utilized a pre-trained image encoder to project all of the frames to the latent space. While most of the methods operate in $\mathcal{W}+$ space~\cite{abdal2019image2styleganpp}, we opt to use both $w^+$ and deep features~\cite{feature-style,parmar2022sam,yin2022styleheat} of the pre-trained StyleGAN to accurately reconstruct the original image and synthesize motion.

%% file: source/400_method.tex
\begin{figure*}
  \centering
  \includegraphics[width=\linewidth]{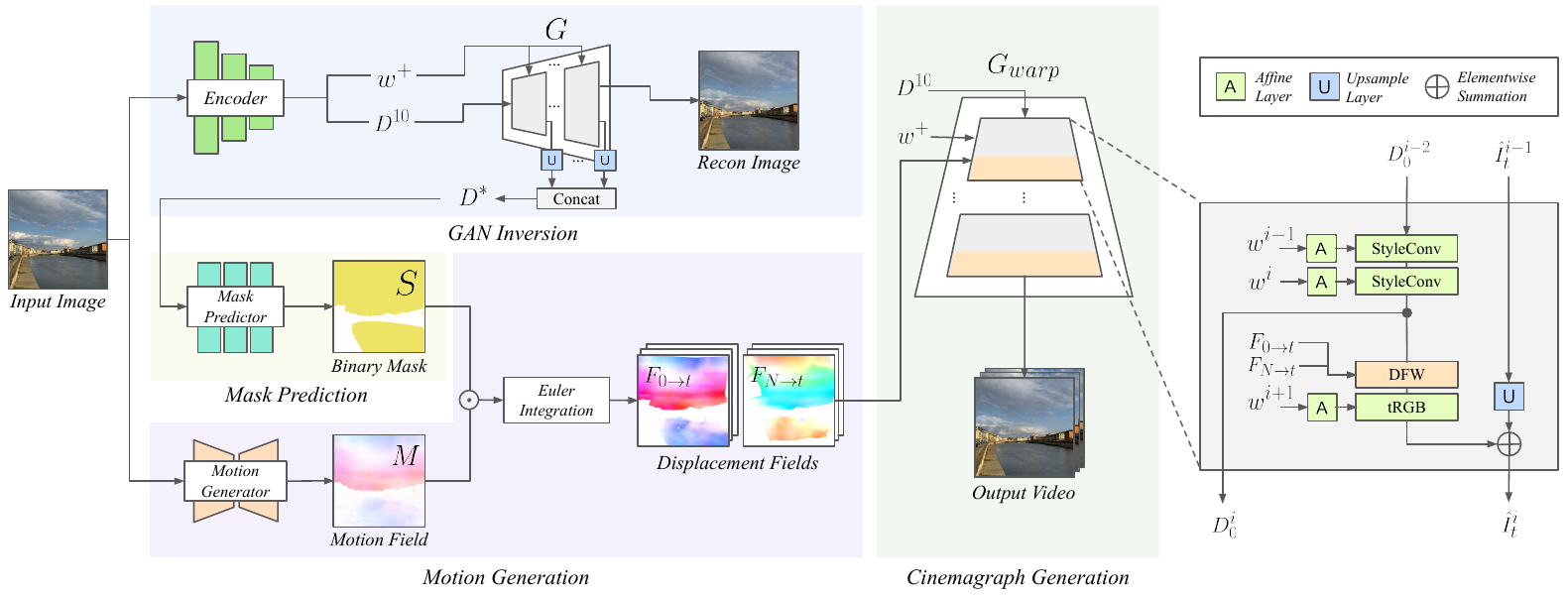}
  \vspace{-0.7cm}
  \caption{\textbf{Overview of StyleCineGAN.} Given an input landscape image $I$, our goal is to generate a cienemagraph using a fixed pre-trained StyleGAN $G$. We project the image into both latent codes $w^+$ and deep features $D^{10}$ of $G$. Using the deep features $D^*$, a mask predictor predicts a segmentation mask $S$. To animate the input image, we use a motion generator to predict the motion field $M$ from $I$. $M$ is refined using $S$. Through Euler integration, $M$ produces the future and past displacement fields $F_{0\to t}$ and $F_{N\to t}$. 
  To synthesize cinemagraph frames, we add a DFW layer in between the layers of $G$. DFW refers to Eqns.~\ref{eqn:dfw1} and ~\ref{eqn:dfw2}. This modification enables the intermediate features of $G$ to be warped according to $F_{0\to t}$ and $F_{N\to t}$ using a joint splatting method at different resolutions, specifically for the StyleGAN layers indexed with $i\in [10, 12, 14, 16, 18]$. The warped deep features are used to synthesize frames $\hat{I}_t$ resulting in the final cinemagraph video.} 
  \vspace{-0.4cm}
  \label{overview} 
\end{figure*}


\section{Methods}
An overview of our method is shown in Figure~\ref{overview}. Given a landscape image $I$, our method generates a seamlessly looping cinemagraph $V=\{\hat{I}_0,...,\hat{I}_{N}\}$
using a pre-trained StyleGAN. At the core of our method, we use a pre-trained generator without additional fine-tuning.
The overall process~is~described~below.

\begin{itemize}
    \item[(A)] First, we project the landscape image into both the latent space and the feature space of StyleGAN. To this end, we train an encoder that outputs both latent codes $w^+$ and intermediate features $D^{10}$ of StyleGAN (Sec.~\ref{method:gan_inv}).
    \item[(B)] In addition, we predict a mask $S$ to divide the image into static and dynamic regions (Sec.~\ref{method:mask_predict}).
    \item[(C)] Next, we use a motion generator that accepts the landscape image $I$ to synthesize a motion field $M$, which defines the position of each pixel in the future frames (Sec.~\ref{method:motion_gen}).
    \item[(D)] Lastly, we generate the final cinemagraph frames using the pre-trained StyleGAN with deep feature warping (DFW) operation added in between the original layers. (Sec.~\ref{method:cinemagraph_gen}).
\end{itemize}
In the following sections, we will describe the details of each process.

\input{source/410_gan_inversion}

\input{source/420_mask}

\input{source/430_motion}

%% file: source/410_gan_inversion.tex
\subsection{GAN Inversion}\label{method:gan_inv}
The first step for generating a cinemagraph is to project the input image to the latent space of a pre-trained StyleGAN.
This process is necessary because StyleGAN is an unconditional generator that cannot take a conditioning input. Many previous methods have used $\mathcal{W}+$~\cite{abdal2019image2styleganpp} latent space, which is an extended space of native latent space $\mathcal{W}$. However, we observed that $w^+\in\mathcal{W}+$ is not expressive enough to reconstruct the original high-frequency details of the landscape images. Therefore, we chose to use the generated deep features $D^{10}$ of StyleGAN as well as $w^+$. The use of $D^{10}$ enables us to recover details in the original input as shown in Figure~\ref{ablation_inversion_fig}.



To project an image $I$ to both the latent space and feature space and obtain $w^+$ and $D^{10}$, we train an encoder $E$ similarly to ~\citet{feature-style}, in which the encoder takes $I$ as input and predicts $(w^+, D^{10})$. Additional training details can be found in the supplementary material.

%% file: source/420_mask.tex
\subsection{Mask Prediction}\label{method:mask_predict}
We improve the quality of GAN inversion described in Sec.~\ref{method:gan_inv} by training an additional classifier that predicts a mask to separate the static and dynamic regions. Using the mask, the structure of the static regions can be preserved. To this end, we train a multi-layer-perceptron (MLP) classifier that accepts the deep features of StyleGAN as its input and outputs the mask that specifies each dynamic region in the image, as performed in DatasetGAN~\cite{zhang2021datasetgan}.


To train the classifier, we manually annotate 32 segmentation masks $S$ of the selected images. We then follow the same procedure as DatasetGAN but using the deep features $D^i$ where $i \in \{10, 11, ..., 18\}$.
We resize all these deep features followed by concatenating them in the channel dimension to construct the input feature $D^{*}$. In the end, paired data $(D^{*}, S)$ is constructed. An ensemble of 10 MLP classifiers is trained with a cross-entropy loss.
To further improve the performance of the previous work, at inference, we also refine the predicted mask. For additional details, please refer to the supplementary material.

%% file: source/430_motion.tex
\subsection{Motion Generation}\label{method:motion_gen}
To generate motion from a static landscape image $I$, we use an Eulerian motion field $M$ which represents the immediate velocity of each pixel.
To predict $M$ given $I$, an image-to-image translation network~\cite{pix2pixHD} is trained as a motion generator. We train the network with paired data $(I, M)$ as performed in the previous method~\cite{Holynski_2021_CVPR}. Adding controls in motion generation is also possible, either by using text prompts~\cite{text2cinemagraph} or drawing arrows~\cite{controllable}.

While the trained network works well to a certain degree, we observed that the predicted motion field $M$ does not align with the boundaries between the static and dynamic regions and often contains motion in static regions. Thus, we further refine $M$ by multiplying it with the segmentation mask $S$, which effectively removes most errors. 
The refined $M$ is then used to simulate the motion of a pixel at time $t$ using the Euler integration as follows:
\begin{align}
    &x_ {t+1} = x_{t} + M(x_{t}), \quad M(x_{t}) = F_{t\to t+1}(x_t), \notag\\
    & F_{0\to t}(x_0) = F_{0 \to t-1}(x_0) + M(x_0 + F_{0 \to t-1}(x_0)),
\end{align}
where $x_0$ is the source pixel position at time 0, $F_{0\to t}$ is the displacement field that defines the displacement of the pixel position $x_0$ to the pixel position $x_t$ at time $t$.

\subsection{Cinemagraph Generation}\label{method:cinemagraph_gen}
After acquiring the displacement field $F_{0\to t}$ (Sec.~\ref{method:motion_gen}), latent codes $w^+$, and deep features $D^{10}$ of StyleGAN (Sec.~\ref{method:gan_inv}), we feed these data into the pre-trained StyleGAN with our DFW layer added. We apply forward warping (or splatting) to the original content to generate a cinemagraph. Here, the application of forward warping directly to an RGB image often results in images with tearing artifacts.
To reduce the artifacts, we warp the deep features $D^{i}_{0}$ of StyleGAN in different resolutions or scales:
\vspace{-0.1cm}
\begin{equation}
    D^{i}_{t} = Warp(D^{i}_{0}, F_{0\to t}),
    \label{eqn:dfw1}
\end{equation}
where $Warp$ is the forward warping function and $D^{i}_{t}$ is the warped deep feature at time $t$ and scale $i$. We observed that warping a single deep feature (e.g., only $D^{10}_{0}$) results in blurry textures in the generated cinemagraphs.~Therefore, we opt to warp the multi-scale deep features; we call this operation a MSDFW. Specifically, we apply warping to the deep features $D^{i}_{0}$ where $i\in [10, 12, 14, 16, 18]$. The deep feature $D^{10}_{0}$ is acquired using GAN inversion, and the consequent deep features $D^{12}_{0}, D^{14}_{0}, ...,$ and $D^{18}_{0}$ are subsequently generated using both $D^{10}_{0}$ and $w^+$, as shown in the rightmost column of Figure~\ref{overview}.



Looping video $V$ with frame length $N+1$ is synthesized using the joint splatting technique~\cite{Holynski_2021_CVPR} using the future and past displacement fields $F_{0\to t}$ and $F_{N\to t}$, generated through Euler integration. The displacement fields are computed with $t$ times of integration on $M$ for $F_{0\to t}$, and $N-t$ times of integration on $-M$ for $F_{N\to t}$. The multi-scale feature $D^{i}_{0}$ is warped in both directions and is composited to form $D^{i}_{t}$. Specifically, $D^{i}_{t}$ is computed as a weighted sum of $Warp(D^{i}_{0}, F_{0\to t})$ and $Warp(D^{i}_{0}, F_{N\to t})$ as follows:
\begin{multline}
    D^{i}_{t}(x') = \frac{\sum_{x \in \chi} \alpha_{t} \cdot Warp(D^{i}_0(x), F_{0\to t})}{\sum_{x \in \chi} \alpha_{t}} \\
    + \frac{\sum_{x \in \chi} (1 - \alpha_{t}) \cdot Warp(D^{i}_{0}(x), F_{N\to t})}{\sum_{x \in \chi} (1 - \alpha_t)},
    \label{eqn:dfw2}
\end{multline}
where $x \in \chi$ is a set of pixels being mapped to the same destination pixel $x'$, and $\alpha_{t}$ is the looping weight defined as $(1-\frac{t}{N})$.

With the above DFW module, we can generate a video frame given the predicted deep features $D^{10}$, latent code $w^+$, and mask $S$:
\vspace{-0.2cm}
\begin{multline}
    \hat{I}_t = S\odot(G_{warp}(D^{10}_{0},w^+,F_{0\to t},F_{N\to t})) \\
    + (\mathbf{1}-S) \odot I,
    \label{eq:cinemagraph}
\end{multline}
where $\odot$ is an element-wise multiplication and $G_{warp}$ is a fixed pre-trained StyleGAN that incorporates our DFW module.

\vspace{-0.45cm}

\paragraph{Style Interpolation}
In addition to motion generation, the change of appearance style is an additional feature often observed in a landscape cinemagraph. This can be achieved by interpolating the latent code with the target latent code as follows:
\begin{equation}
    w^+_{s} = w^+ \cdot (1- \beta) + w^+_{t}  \cdot \beta,
\label{eq:interp}
\end{equation}
where $w^+_{s}$ is the interpolated latent code, $w^+_{t}$ is the latent code of the target style, and $\beta$ is the interpolation weight. This is possible because the later layers of the StyleGAN only modify the color of the synthesized image while the original structure is preserved.

With modified Equation~\ref{eq:cinemagraph}, we prevent the visual mismatch between the static and dynamic regions by also reflecting the changes to the static regions as follows:
\begin{align}
    \begin{split}\label{eqn:style1}
        \hat{I}_t ={}& S\odot(G_{warp}(D^{10}_{0},w^+_{s},F_{0\to t},F_{N\to t}))\\
        & + (\mathbf{1}-S) \odot (I + \Delta I_{s}),
    \end{split}\\
    \begin{split}\label{eqn:style2}
        \Delta I_{s} = G(D^{10}_{0},w^+_{s}) - G(D^{10}_{0},w^+),
    \end{split}
\end{align}
where $\Delta I_{s}$ is the color difference between the two images generated from latent codes $w^+_{s}$ and $w^+$.

%% file: source/600_experiment.tex
\begin{figure} \centering
    \setlength{\itemwidth}{6cm}

\resizebox{\columnwidth}{!}{
\begin{tabular}{cccc}
     \animategraphics[width=\itemwidth, autoplay, final, loop, nomouse, method=widget, poster=last]{24}{graphics/results_small/0002268/}{000}{047} &
     \animategraphics[width=\itemwidth, autoplay, final, loop, nomouse, method=widget, poster=last]{24}{graphics/results_small/0002629/}{000}{047} &
     \animategraphics[width=\itemwidth, autoplay, final, loop, nomouse, method=widget, poster=last]{24}{graphics/results_small/0003571_red/}{000}{047} &
     \animategraphics[width=\itemwidth, autoplay, final, loop, nomouse, method=widget, poster=last]{24}{graphics/results_small/0003706_blue/}{000}{047}
\end{tabular}
}

    \caption{Generated cinemagraph results. \textbf{This figure contains video clips, thus consider viewing it using Adobe Reader.} The first two are cinemagraphs without appearance change, and the last two are cinemagraphs with appearance change.}

    \label{fig:results_second}

    \vspace{-0.5cm}
    
\end{figure}

\section{Experiments}

In this section, we compare our method with state-of-the-art landscape cinemagraph generation methods~\cite{animating-landscape, Logacheva_2020_ECCV, Holynski_2021_CVPR, text2cinemagraph} (Sec.~\ref{subsec:comparison_cine}) and with an unconditional video generation method~\cite{tian2021a} (Sec.~\ref{subsec:comparison_gen}). In addition, we show the importance of the components used in our method through ablation studies (Sec.~\ref{subsec:ablation}). To observe more various results, we recommend readers see the supplementary video. Example results are presented in Figures~\ref{fig:teaser} and ~\ref{fig:results_second}.


\begin{figure}[t!]
    \centering
    \includegraphics[width=.9\columnwidth]{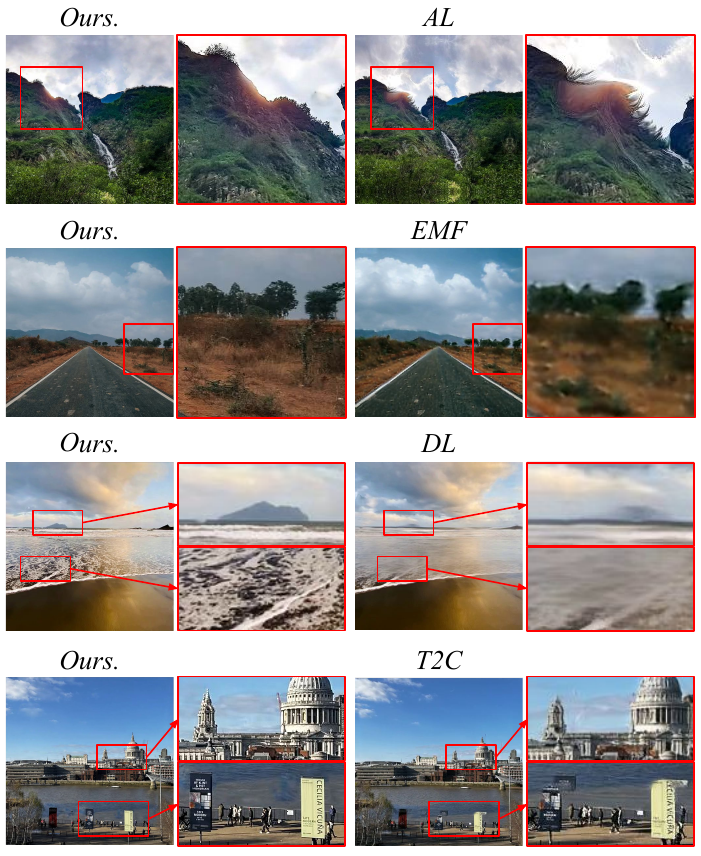}
    \vspace{-0.3cm}
    \caption{Qualitative comparison with state-of-the-art cinemagraph generation methods.
    Please refer to the supplementary video for more examples.
    }
    \vspace{-0.3cm}
    \label{result_fig_Compare}
\end{figure}


\subsection{Comparisons with Cinemagraph Generation Methods}\label{subsec:comparison_cine}

We compared our method with state-of-the-art cinemagraph generation methods: Animating Landscape (AL)~\cite{animating-landscape}, Deep Landscape (DL)~\cite{Logacheva_2020_ECCV}, Eulerian Motion Field (EMF)~\cite{Holynski_2021_CVPR}, and Text2Cinemagraph (T2C)~\cite{text2cinemagraph}. AL is a learning-based method that uses backward warping to generate cinemagraph frames. DL trains a style-based generator to synthesize images and dynamic motion in a scene. EMF and T2C train an encoder-decoder architecture paired with a motion generator to produce looping cinemagraphs.
We used official implementations of AL, DL, and T2C, and faithfully reproduced EMF based on the provided training details and hyper-parameters.
In the following, we will examine the qualitative difference, evaluate the results using two metrics, and report the results of human perceptual studies.

\vspace{-0.3cm}

\paragraph{Qualitative Comparisons}
Figure~\ref{result_fig_Compare} shows visual results from AL, EMF, DL, T2C, and our method.
AL exhibits stretching artifacts (row 1) due to recurrent backward warping, whereas our method uses forward warping that prevents the appearance of such distortions. EMF struggles with precise reconstruction (row 2), because its encoder-decoder architecture is trained from scratch on a low-quality video dataset. In contrast, our method leverages a StyleGAN model pre-trained on high-quality images, resulting in cinemagraphs with high perceptual quality. DL often loses fine details in textures (row 3) as it performs GAN inversion using the latent codes of StyleGAN, and fine-tunes the generator. Our method, utilizing deep features for GAN inversion, retains original texture details. T2C encounters difficulties in preserving high-frequency details, and accurate motion segmentation between static and dynamic regions (row 4). In contrast, our method utilizes a pre-trained StyleGAN for cinemagraph generation, and uses its features for mask prediction, preserving both fine and structural details of the source image.

\begin{table}[t!]\centering
\caption{Quantitative comparison with the state-of-the-art cinemagraph generation methods. We compared our method with AL, EMF, DL, and T2C.
\vspace{-0.2cm}
The best scores are bolded.}
\resizebox{.95\columnwidth}{!}{
\begin{tabular}{@{\extracolsep{4pt}}cccccccc@{}}
\hline
         & \multicolumn{3}{c}{Static Consistency}                   & \multicolumn{4}{c}{Motion Quality}\\ \cline{2-4} \cline{5-8} \cline{8-8}
Method   & LPIPS$\downarrow$ & MS-SSIM$\uparrow$ & RMSE$\downarrow$ & LPIPS$\downarrow$ & MS-SSIM$\uparrow$ & RMSE$\downarrow$  & FID$\downarrow$ \\ \hline
AL       & 0.0477            & 0.9524            & 8.8956           & 0.0617            & 0.6819            & 25.1089           & 53.6893  \\
EMF      & 0.0103            & 0.9723            & 6.0440           & 0.0533            & 0.7102            & 22.8932           & 45.8035  \\
DL       & 0.0071            & 0.9931            & 2.1044           & 0.0504            & 0.7062            & 21.8011           & 41.6374  \\
T2C      & 0.0063            & 0.9773            & 4.8785           & 0.0159            & 0.7224            & 21.2784           & 40.1186         \\
Ours     & \textbf{0.0062}   & \textbf{0.9962}   & \textbf{1.9430}  & \textbf{0.0131}   & \textbf{0.7237}   & \textbf{20.8948}  & \textbf{39.2113}  \\ \hline
\end{tabular}}
\vspace{-0.4cm}
\label{table_PC}
\end{table}

\vspace{-0.3cm}

\paragraph{Quantitative Comparisons}

The evaluation was performed considering two aspects: (1) static consistency and (2) motion quality.
Static consistency measures the consistency over time of non-movable regions such as buildings and mountains.  
Motion quality refers to both the image quality of the animated regions and the animation plausibility.
For a fair comparison, we normalized the speed of motion in the generated results for all methods to match the average speed in the test dataset.
AL, EMF, DL, and T2C were provided with the first frame $I_0$ of the test video to generate 512$\times$512 videos. Our method generates 1024$\times$1024 results, thus we downsampled it to the size of 512$\times$512.

For a quantitative evaluation, we used 224 test videos from the Sky Time-Lapse dataset~\cite{Xiong_2018_CVPR}. To measure the static consistency, we computed LPIPS~\cite{zhang2018perceptual}, Root-Mean-Squared Error (RMSE), and MS-SSIM~\cite{Wang2003MultiscaleSS} between generated frame $\hat{I}_n$ and the input image $I_0$ with the sky masked out. To measure the motion quality, we used the same evaluation metrics along with Fréchet inception distance (FID)~\cite{Seitzer2020FID} between $\hat{I}_n$ and the ground-truth frame $I_n$ with the static parts masked out.

Table~\ref{table_PC} reveals the quantitative evaluation results.
For both static consistency and motion quality, our method outperforms the other approaches.
Our cinemagraphs represent static consistency better because use of the mask improves the accuracy in the detection of static regions.
In addition, the quality of texture details are improved in the generated frames due to GAN inversion and the MSDFW based on the deep features of StyleGAN.
The use of our motion generator and DFW leads to improved motion quality for each scene, compared with that of the results of previous approaches.


\vspace{-0.3cm}

\paragraph{User Study}
We conducted a user study with 17 participants to subjectively evaluate our method in comparison with previous cinemagraph generation methods. We did not target any specific demographic groups, as the evaluation of cinemagraphs should not be dependent on particular demographic distributions. We conducted two evaluations: score evaluation and side-by-side comparison. Both evaluations assessed static consistency and motion quality. In the score evaluation, participants rated the video presented with a reference image on a 1-to-5 scale, with "very bad" being 1 and "very good" being 5. In the side-by-side comparison, participants selected the preferred cinemagraph from two videos generated using different methods, presented with a reference image. For both evaluations, we used ten samples randomly chosen from the LHQ~\cite{LHQ} dataset. To eliminate bias, distinct samples were used for each evaluation, and the positions of the cinemagraphs were randomized in all questions.









\begin{table}[t!]\centering
\caption{Human perceptual study for assessing the overall quality. The participants were asked to score the given cinemagraphs based on two criteria.
The best scores are bolded.
}
\vspace{-0.2cm}
\resizebox{.80\columnwidth}{!}{
\begin{tabular}{@{}ccc@{}}
\hline
Method & Static Consistency       & Motion Quality            \\ \hline
AL     & 1.59 $\pm$ 0.20          & 1.76 $\pm$ 0.45           \\
EMF    & 2.35 $\pm$ 0.89          & 2.24 $\pm$ 0.99           \\
DL     & 3.35 $\pm$ 0.83          & 3.25 $\pm$ 0.67           \\
T2C    & 3.75 $\pm$ 0.69          & 2.96 $\pm$ 0.66           \\
Ours   & \textbf{4.37 $\pm$ 0.18} & \textbf{3.86 $\pm$ 0.53}  \\ \hline
\end{tabular}}
\label{table_user_study_score}
\end{table}

\begin{table}[t!]\centering
\caption{Human perceptual study with side-by-side comparison. The participants were asked which of the two cinemagraphs were better in terms of static consistency and motion quality. The percentage of ours chosen against each competing method is reported.}
\vspace{-0.2cm}
\resizebox{.80\columnwidth}{!}{
\begin{tabular}{@{}ccc@{}}
\hline
               & \multicolumn{2}{c}{Human Preference (Ours \%)}          \\ \cline{2-3}
Method         & Static Consistency    & Motion Quality                  \\ \hline
$vs$ AL        & \textbf{99.35}\%      & \textbf{92.81}\%                \\
$vs$ EMF       & \textbf{96.08}\%      & \textbf{86.93}\%                \\
$vs$ DL        & \textbf{94.77}\%      & \textbf{84.97}\%                \\
$vs$ T2C       & \textbf{77.78}\%      & \textbf{73.86}\%                \\\hline
\end{tabular}}
\label{table_user_study_pref}
\vspace{-0.4cm}
\end{table}

Tables~\ref{table_user_study_score} and \ref{table_user_study_pref} summarize the statistics of the user studies. For both score evaluation and human preference, our method outperforms the previous approaches by a substantial margin.
We also performed statistical analysis (Kruskal-Wallis H-test) on the resulting evaluation scores. The results showed that our method achieved significantly higher scores in all pairs of comparisons with every previous method (p$<$0.001 in post-hoc analysis).
Both user study results reveal the advantages of our method for generating cinemagraphs with high image quality and plausible animation.

\vspace{0.1cm}

\subsection{Comparisons with Video Generation Methods}
\label{subsec:comparison_gen}

We compared our approach with MoCoGAN-HD~\cite{tian2021a}, a video generation method that predicts the trajectory within the latent space of a pre-trained StyleGAN. For content generation, we used the same pre-trained StyleGAN as that used by MoCoGAN-HD, which was trained on the Sky Time-lapse~\cite{Xiong_2018_CVPR} dataset to generate 128$\times$128 images. The comparison was made in an unconditional manner using 300 randomly sampled latent codes, at a resolution of 128$\times$128. We compared the first 60 frames of the generated videos with those of MoCoGAN-HD, as our method produces looping videos.  


\begin{figure}[htb!]
    \centering
    \includegraphics[width=.88\columnwidth]{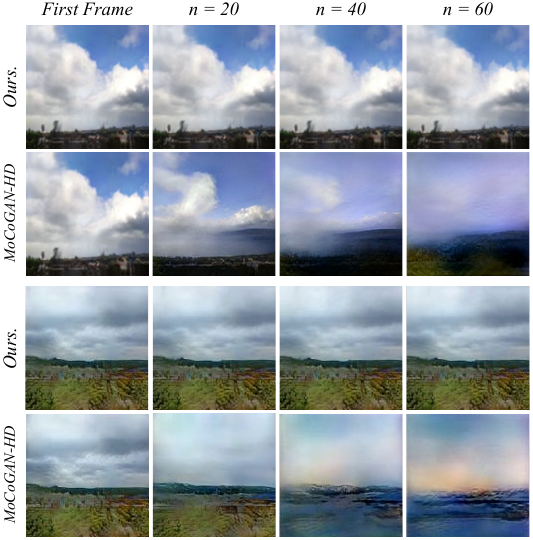}
    \vspace{-0.2cm}
    \caption{
    Qualitative comparison with the state-of-the-art video generation method, MoCoGAN-HD~\cite{tian2021a}. For more examples from this comparison, please refer to the supplementary video.
    }
    \vspace{-0.25cm}
    \label{result_fig_SC}
\end{figure}

\begin{figure}[htb!]
    \centering
    \includegraphics[width=0.9\columnwidth]{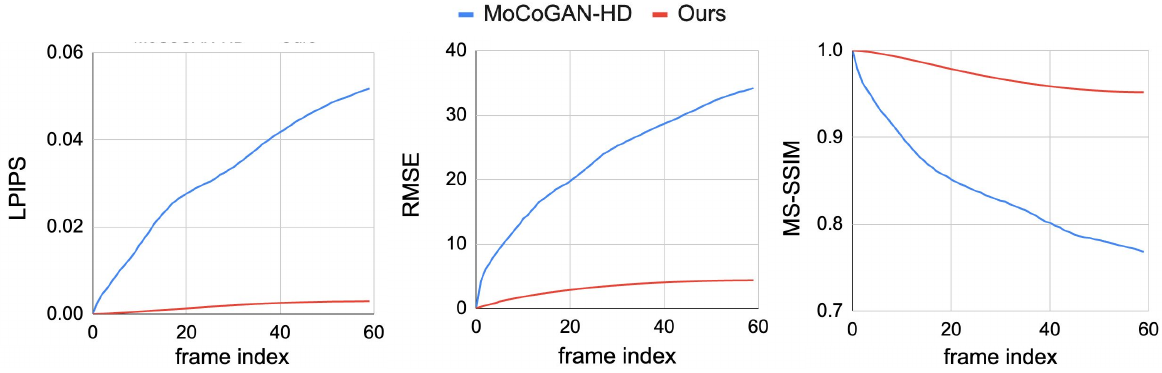}
    \vspace{-0.2cm}
    \caption{Quantitative comparison of content preservation with state-of-the-art video generation method MoCoGAN-HD~\cite{tian2021a}.
    }
    \label{graph_SC_quan}
    \vspace{-0.5cm}
\end{figure}

\vspace{-0.35cm}

\paragraph{Qualitative Comparisons}
Figure~\ref{result_fig_SC} presents a qualitative comparison. As shown in the first and third rows, the videos generated using our method exhibit static consistencies, with the clouds moving while the ground remains static. In contrast, as shown in the second and fourth rows, the frames generated by MoCoGAN-HD deviate significantly from the original image, making the method unsuitable for cinemagraph generation.





\vspace{-0.35cm}

\paragraph{Quantitative Comparisons}
We compared the content preservation ability of both methods by measuring LPIPS~\cite{zhang2018perceptual}, RMSE, and MS-SSIM~\cite{Wang2003MultiscaleSS} between the first frame $\hat{I_0}$ and the subsequent frames $\hat{I}_n$. Because the main content to be preserved in cinemagraphs is the static regions, we defined content preservation as static consistency. We masked out the sky for all image frames according to the segmentation mask and compared only the static parts. We used 219 videos in which the scene could be horizontally divided into static and animated regions.
LPIPS, RMSE, and MS-SSIM were all computed for each pair of $\hat{I_0}$ and $\hat{I}_n$, and were averaged over the number of samples. Figure~\ref{graph_SC_quan} shows that MoCoGAN-HD exhibits significant divergence in terms of LPIPS, RMSE, and MS-SSIM over time. This indicates that MoCoGAN-HD is unable to generate motion that preserves the content of the original image. In contrast, our method exhibits a small diverging trend, which confirms the superiority of our method and its ability to preserve the content of the original image over time, by utilizing deep features.



\subsection{Ablation Study}
\label{subsec:ablation}
\vspace{-0.1cm}
To demonstrate the effectiveness of our design choices for the proposed method, we conducted a series of ablation studies. Specifically, we focused on the effectiveness of our warping and GAN inversion methods. To evaluate warping, we compared the image frames generated with and without the use of 1) forward warping, 2) DFW, 3) MSDFW, and 4) segmentation mask. The first frames of 224 test videos from the Sky Time-Lapse~\cite{Xiong_2018_CVPR} dataset were given as input to generate 1024$\times$1024 videos, and we used the first 60 frames for the evaluation. For the evaluation of GAN inversion, we compared images reconstructed with and without the use of the deep features. A total of 256 images, 128 from the Sky Time-Lapse and 128 from the Eulerian~\cite{Holynski_2021_CVPR} dataset were
provided
to generate 1024$\times$1024 reconstructed images.

\vspace{-0.2cm}

\begin{figure}
    \centering
    \includegraphics[width=.95\columnwidth]{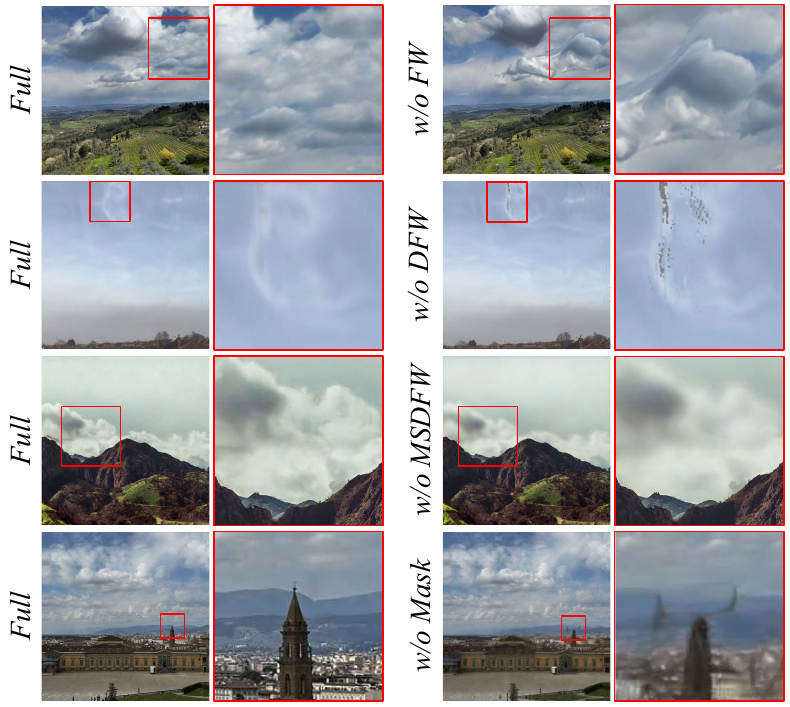}
    \vspace{-0.2cm}
    \caption{
    Results of qualitative comparison in ablation study. Please see the suuplmentary video for the animated results of this evaluation.
    }
    \vspace{-0.3cm}
    \label{ablation_fig}
\end{figure}

\begin{figure}
    \centering
    \includegraphics[width=.9\columnwidth]{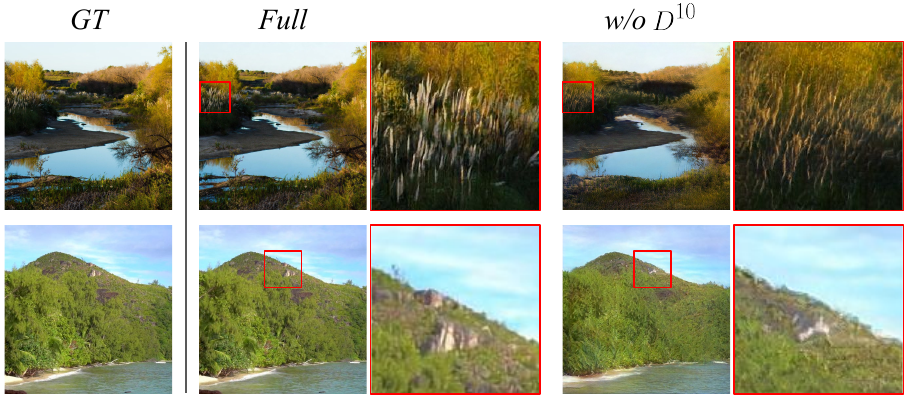}
    \vspace{-0.2cm}
    \caption{
    Effect of deep feature inversion.
    Excluding deep feature $D^{10}$ during GAN Inversion resulted in an inaccurate reconstruction of the original image.}
    \vspace{-0.5cm}
    \label{ablation_inversion_fig}
\end{figure}

\vspace{-0.cm}
\paragraph{Forward Warping}\label{forward warping}
We compared the image frames generated with and without forward warping (FW). For the case without FW, we warped the deep features of StyleGAN using backward warping. The first row in Figure~\ref{ablation_fig} shows the qualitative results of this comparison. As revealed in the figure, without FW, the results usually contain unrealistically stretched textures. For quantitative comparisons, we computed LPIPS, MS-SSIM, and RMSE between $I_n$ and $\hat{I}_n$. Table~\ref{table_ablation} shows that excluding forward warping resulted in significant degradation in terms of MS-SSIM. This indicates that the stretched textures resulted in huge structural distortions in the generated images.


\begin{table}[]\centering
\caption{
Results of quantitative evaluation in ablation study.
Each component in our warping method increased the perceptual quality of the results. The best scores are bolded.}
\vspace{-0.2cm}
\resizebox{.80\columnwidth}{!}{
\begin{tabular}{@{}lccc@{}}
\hline\hline
Method           & LPIPS$\downarrow$ & MS-SSIM$\uparrow$ & RMSE$\downarrow$  \\ \hline
Ours - Full      & \textbf{0.0511}   & \textbf{0.7165}   & \textbf{21.8188}  \\ \hline
Ours - w/o FW    & 0.0524            & 0.6853            & 22.9816           \\
Ours - w/o DFW   & 0.0537            & 0.6908            & 22.4993           \\
Ours - w/o MSDFW & 0.0629            & 0.6946            & 22.3227           \\
Ours - w/o Mask  & 0.0564            & 0.6980            & 22.6213           \\ \hline\hline
\end{tabular}}
\vspace{-0.2cm}
\label{table_ablation}
\end{table}

\begin{table}[]
\centering
\caption{Quantitative evaluation results on GAN inversion. The best scores are bolded.
}
\vspace{-0.2cm}
\resizebox{.80\columnwidth}{!}{\begin{tabular}{cccccc}
\hline
$w^{+}$ & $D^{10}$ & FID$\downarrow$ & LPIPS$\downarrow$  & RMSE$\downarrow$  & MS-SSIM$\uparrow$       \\ \hline
$\checkmark$ & --                   & 40.8276            & 0.0153            & 18.5901            & 0.7891          \\
$\checkmark$ & $\checkmark$         & \textbf{11.8231}   & \textbf{0.0019}   & \textbf{5.6410}    & \textbf{0.9907} \\
\hline
\end{tabular}}
\vspace{-0.4cm}
\label{table_inversion}
\end{table}

\vspace{-0.3cm}

\paragraph{Deep Feature Warping}\label{feature warping}
We then compared frames generated with and without DFW. For the case without DFW, we directly warped RGB images using predicted motion fields for comparison. The second row in Figure~\ref{ablation_fig} presents the qualitative comparison. As shown in the figure, the frame generated without DFW contained tearing artifacts in the animated regions.
Table~\ref{table_ablation} shows an increase in the LPIPS score, which indicates that excluding DFW degraded the perceptual quality by introducing tearing artifacts.



\vspace{-0.3cm}

\paragraph{Multi-scale Deep Feature Warping}
Image frames generated with and without MSDFW were also compared. For the case without MSDFW, only a single deep feature $D^{10}$ was warped and propagated through the next blocks of StyleGAN. The results of the qualitative comparison are shown in the third row of Figure~\ref{ablation_fig}. As revealed in the figure, excluding MSDFW resulted in blurry textures in the dynamic region. The quantitative comparison reported in Table~\ref{table_ablation} shows a significant increase in the LPIPS score. This indicates that excluding MSDFW degraded the perceptual quality, especially the texture details.

\vspace{-0.3cm}

\paragraph{Segmentation Mask}
We compared the image frames generated with and without a segmentation mask. For the case without a mask, we used the initially predicted motion field for warping. The fourth row in Figure~\ref{ablation_fig} shows the qualitative result of this comparison. As shown in the figure, excluding the mask resulted in erroneous movement in the static regions. The quantitative comparison reported in Table~\ref{table_ablation} reveals a significant degradation of MS-SSIM, which indicates the structural distortion caused by the erroneous motion.

\vspace{-0.3cm}

\paragraph{Deep Feature Inversion}\label{feature inversion}
To evaluate the effectiveness of GAN inversion, we compared the images reconstructed with and without the use of the deep features $D^{10}$. For the case without deep feature inversion, only the latent codes $w^+$ of a pre-trained StyleGAN were used to reconstruct the input landscape image. Figure~\ref{ablation_inversion_fig} shows a qualitative result of this comparison. As revealed in the figure, using only $w^+$
failed to accurately reconstruct the details of the original images.
For quantitative comparisons, FID, LPIPS, RMSE, and MS-SSIM were computed between $I$ and $\hat{I}$.
Table~\ref{table_inversion} reveals significant improvements in the perceptual quality of the results when reconstructed using the deep features $D^{10}$.

%% file: source/700_discussion.tex
\section{Limitations and Future Work}

\vspace{-0.1cm}

\begin{figure}[t!]
    \centering
    \includegraphics[width=\columnwidth]{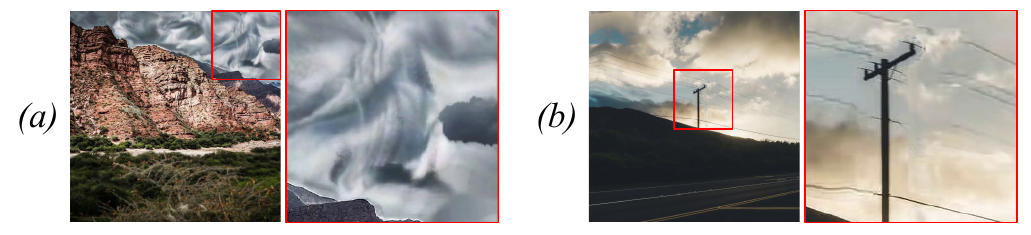}
    \vspace{-0.65cm}
    \caption{Limitations of StyleCineGAN: (a) automatic prediction of motion cannot be accurate for all images, and (b) the motion of a very thin structured object is hard to be isolated.}
    \vspace{-0.5cm}
    \label{fig_limitations}
\end{figure}

Our mask predictor and motion generator generally performed well for landscape images.
However, automatic prediction cannot be accurate for all images because most landscape images contain inherent ambiguity in their motion directions except for some obvious cases (e.g., waterfalls). The failure case is illustrated in Figure~\ref{fig_limitations}~(a).
User-defined motion hints can be used to resolve such ambiguities and provide further control capability during the generation process~\cite{controllable, text2cinemagraph}.
In addition, it is hard for our method to isolate the motion of a very thin structured object placed within the animated region as shown in Figure~\ref{fig_limitations}~(b). This is because our method performs warping of features at multiple resolutions, in which low-resolution features cannot spatially represent the thin structures.

In this work, we mainly focused on animating the landscape images, particularly skies and fluids, while putting other types of animations outside the scope. In future developments, we would like to expand the capabilities to include other forms of motion. Investigating the rotating hands of a clock, the playing arm of a guitarist moving up and down, and a flag or the wings of a bird fluttering for cinemagraph generation would be a very interesting direction to pursue.


%% file: source/800_conclusion.tex
\section{Conclusion}
\vspace{-0.1cm}
We proposed the first approach that leverages a pre-trained StyleGAN for high-quality one-shot landscape cinemagraph generation.
In contrast to previous studies, our method does not require training a large image generator from scratch, and also systematically improves the resolution of the generated cinemagraphs to 1024$\times$1024. At the core of our method, we utilized the deep features of a pre-trained StyleGAN, because those features can help preserve spatial information and encode both high-level semantic and low-level style appearances.
Using our MSDFW approach, we applied the predicted motion to the deep feature space of StyleGAN. Both qualitative and quantitative results confirm that our method substantially outperforms existing baselines.